# Pronunciation Modeling of Foreign Words for Mandarin ASR by Considering the Effect of Language Transfer

*Lei Wang and Rong Tong*

Institute for Infocomm Research, Singapore 138632

{wangl,tongrong}@i2r.a-star.edu.sg

## Abstract

One of the challenges in automatic speech recognition is foreign words recognition. It is observed that a speaker's pronunciation of a foreign word is influenced by his native language knowledge, and such phenomenon is known as the effect of language transfer. This paper focuses on examining the phonetic effect of language transfer in automatic speech recognition. A set of lexical rules is proposed to convert an English word into Mandarin phonetic representation. In this way, a Mandarin lexicon can be augmented by including English words. Hence, the Mandarin ASR system becomes capable to recognize English words without retraining or re-estimation of the acoustic model parameters. Using the lexicon that derived from the proposed rules, the ASR performance of Mandarin English mixed speech is improved without harming the accuracy of Mandarin only speech. The proposed lexical rules are generalized and they can be directly applied to unseen English words.

## 1. Introduction

The performance of automatic speech recognition (ASR) system is often challenged by the variability observed in speakers and speech content. Among the challenges, the task of foreign words recognition attracts researchers' attention as foreign proper names and popular words are widely used in daily conversation. For example, product names such as *iPad* and *iPhone* are frequently used in many languages besides English. It is crucial to recognize such foreign words in the practical applications such as keyword search and dictation.

In this work, the objective of the foreign words recognition task is to use an existing ASR system to recognize the words of different languages occurred in continuous speech. The ASR system is built using speech data of one single language, which is known as native language (L1). The foreign words are from other languages, which are known as target languages (L2). The application scenario of the task assumes that the continuous speech of a native speaker contains a small number of words in L2.

The above mentioned task can be seen as a special case of out-of-vocabulary (OOV) problem. However, foreign words recognition is more challenging than recognizing OOV words of the same language as i) L2 words usually originate from a phonetic system which is different from L1; ii) Higher variability is observed in L2 words' pronunciations as the pronunciations are greatly impacted by the language background of the L1 speaker [1]. If the speaker has limited knowledge of L2 or low proficiency, he may speak L2 in a nativized way [2]. Hence, the existing work on OOV cannot be directly applied to recognize foreign words.

The problem of foreign words recognition is also referred as code-mixing [3]. The existing studies can be grouped into 2 categories: acoustic modeling and pronunciation modeling. Acoustic modeling is to incorporate foreign acoustic characteristics into L1 acoustic models. In [4], English phoneme models are combined with German acoustic model inventory to recognize the English words that occurred in Germany speech. A bilingual acoustic model is proposed in [5] for recognizing both English and Mandarin words. A small amount of non-native data is utilized in [6] to adapt the acoustic model for better accuracy on non-native speech recognition. These methods usually require a certain amount of transcribed training data in order to train or adapt the acoustic models.

Pronunciation modeling in foreign words recognition is to augment the original L1 lexicon with foreign words. Various pronunciation modeling techniques can be categorized into 2 groups: data driven approach and rule-based approach. In data driven approach, the pronunciations of new words are automatically derived from the training data according to a small seed lexicon [7]. In another attempt, a pronunciation mixture model is proposed in [8] to learn stochastic lexicons from speech data. Moreover, pronunciations of foreign words are learned by performing forced alignment using L1 acoustic model in [9]. Besides the automatic techniques in data driven approach, rule-based methods are also explored in modeling foreign accented speech variants [10]. Syllable-based pronunciation variation rules are generated in [11] to improve the recognition performance. Letter to sound rules are also adopted in [12] to derive pronunciations for foreign words. Various methods improves the performance of the foreign word recognition, however the effect of language transfer is not fully exploited in the existing rule-based pronunciation modeling methods.

This paper proposes a rule-based approach for foreign words pronunciation modeling by studying the effect of language transfer [13, 14, 15, 16]. When a speaker speaks a non-native language (L2), he often applies his knowledge of the native language (L1) to L2. Such phenomena is referred as the effect of language transfer. The language transfer effect is observed when a native Mandarin speaker pronounces English words. For instance, an English word can end with a consonant, while Mandarin words usually end with a vowel or a nasal sound. Hence, when a native Mandarin speaker pronounces an English word which ends with a consonant sound, he may append a vowel sound to the last consonant, e.g. 'book /b u k/' could be mispronounced as /b u k ə/. Based on the study of language transfer, we propose a set of lexical rules to represent English words' pronunciations using Mandarin phoneme sequences.

The proposed rule-based approach has the following advantages: i) There is no need to retrain acoustic models or re-estimate acoustic parameters. ii) Once the set of rules is generated, the pronunciations of unseen words can be derived automatically. iii) The set of rules can be easily updated without modifying acoustic model and LM.

The remaining portion of the paper is organized as follows: Section 2 introduces the concept of the effect of language transfer. The proposed lexical rules for mapping English words' pronunciations to Mandarin phoneme sequences are also presented. Section 3 describes the experimental setup. The experimental results and observations are reported in Section 4. Finally, conclusion and future work are given in Section 5.

## 2. Language Transfer

Language transfer is a psychological phenomenon observed in foreign language acquisition. It refers to the influence resulting from the similarities and differences between the target language (L2), and any other language that has been previously acquired, e.g. mother language (L1) [14, 15, 16]. According to its effects, language transfer can be categorized into positive transfer and negative transfer. Positive transfer occurs when similar language characteristics are shared between L1 and L2, learners can easily apply his knowledge of L1 on L2. For example, the English phoneme /a:/ (e.g. *bar*) is close to Mandarin vowel 'a' (e.g. *ba1*) in articulation, hence it is straightforward for a native Mandarin speaker to pronounce /a:/ properly in English.

Negative transfer occurs when L1 and L2 have different interpretations on certain language characteristics. Under such scenario, the learner may make errors if he applies the L1 rules on L2. For instance, /ʌ/ (e.g. *cut*) sound does not exist in Mandarin phonological system. A native Mandarin speaker may mispronounce /ʌ/ with 'a' (or /a:/).

In this work, we focus on the ASR of Mandarin speech mixed with English words. When a native Mandarin speaker speaks English words (e.g. *iPhone*, *Windows*), his pronunciations of the English words could be impacted by the effects of language transfer from his Chinese knowledge. Inspired by the effect of negative transfer, this work explores a way to recognize English words using the existing Mandarin ASR system without updating the acoustic model. We propose to model native Mandarin speakers's English pronunciations by using a set of rules. These rules are derived by considering the effects of Chinese to English language transfer.

Different levels of language transfer are observed in the study of second language acquisition, for example, phonetic, grammatical, lexical and pragmatic level. In this paper, we focus on phonetic aspect of language transfer in English pronunciation modeling.

### 2.1. Phonology difference in Mandarin and English

Mandarin Chinese and English belong to different language families. English is a Germanic language while Mandarin Chinese is from the Sino-Tibetan language family. They have different phonological structures, and Table 1 summarizes the major phonological differences between Mandarin and English [17].

Mandarin is a tonal, monosyllabic language while English is a multisyllabic language without tones. Mandarin does not allow consonant cluster and each syllable follows the structure of (C)(G)V(N), where C denotes a consonant, G denotes a glide sound, V denotes a vowel and N denotes a nasal. English allows consonants in a cluster with the structure of (C)(C)(C)V(C)(C)(C)(C). It also allows multiple numbers of consonants as word final (except /h/) while Mandarin generally does not allow consonants as syllable final.

Table 1: *Phonological differences between Mandarin Chinese and English.*

| Phonological structure | Mandarin | English |
|---|---|---|
| Words | Monosyllabic | Multisyllabic |
| Tone | 5 tones (including neutral tone) | No tone |
| Syllable structure | (C)(G)V(N) e.g. *chuang* | (C)(C)(C)V(C)(C)(C)(C) e.g. *strengths* |
| Syllables | Consonant clusters are not allowed | Clusters of consonants, e.g., *establish* |
| Final consonant | Generally not allowed | Allowed except /h/, e.g. *mass*, *pad* |

In ASR, Mandarin Chinese and English are also different in numbers of phonemes. In CMU dictionary [18], 39 English phonemes are defined, excluding silence model. In Mandarin ASR, different sets of phonemes were examined for acoustic modeling. For example, in [13], 64 toneless initial-final units (excluding silence and garbage models) were used to build ASR system. In another system [19], researchers defined 36 toneless monophones hence to reduce the complexity of the acoustic model. Due to the differences between Mandarin and English phonological systems, some English phonemes do not appear in the Mandarin, e.g. sound /θ/ (or /TH/ in CMU dictionary) in word *thin* does not exist in Mandarin phonological system. Hence, to assign a Mandarin phoneme sequence to an English word, a close Mandarin phoneme should be selected to substitute the corresponding English phoneme.

In the next section, a few guidelines are described to map each English phoneme to a Mandarin phoneme.

### 2.2. General phoneme mapping

One of the simplest ways to assign Mandarin phoneme sequences to English words is mapping each English phoneme to the closest Mandarin phoneme [20]. To identify the closest phoneme, the following criteria are used as guidelines: i) Both the English phoneme and its corresponding Mandarin phoneme should be from the same phonetic groups, e.g. vowel, consonant and nasal. ii) When pronouncing the phonemes, the mouth and tongue positions should be similar. Table 2 shows the phonemes mapping table from English to Mandarin. The English phonemes definition follows the CMU dictionary [18], and the corresponding Mandarin phonemes are shown as basic units in Chinese Pinyin.

### 2.3. Mapping rules by considering L1 transfer

In Table 2, many of phonemes in the 4 consonant categories and nasal sounds are common in both English and Mandarin, hence such English phonemes are mapped to the corresponding Chinese Pinyin syllables/phonemes. However, due to the effect of negative language transfer, 2 issues can be observed on i) consonants not followed by vowels, and ii) nasal sounds not followed by vowels. The next sections discuss the 2 issues in details, and the English phoneme notations in the below examples follow the CMU definition [18] instead of IPA.

Table 2: *Direct mapping from English phonemes to Chinese Pinyin.*

| English phonemes as defined in the CMU dictionary | Chinese Pinyin syllables | Categories |
|---|---|---|
| AA, AE, AH, AO, AW, AY, EH, ER, EY, OY | ao, ai, a, ao, ao, ai, ai, e, ei, ao | open-mouths vowels |
| IH, IY | i, i | even-teeth vowels |
| OW, UH, UW | ou, u, u | close-mouths vowels |
| B, D, G, P, T, K | b, d, g, p, t, k | plosive consonants |
| F, S, SH, TH, R, HH | f, s, x, s, r, h | fricative consonants |
| Z, CH, DH, ZH, JH | z, q, zh, zh, j | affricate consonants |
| M, N, NG | m, n, ng | nasal |
| L | l | lateral consonants |
| V, W, Y | w, w, y | approximants |

*2.3.1. Consonants not followed by vowels*

In Chinese Pinyin, consonants are usually followed by vowels, however, English does not have such constraints. When a native Mandarin speaker pronounces the English words which contains consecutive consonants, i.e. in the form of (C)(C)(C)V(C)(C)(C)(C), he may insert vowel after each consonant. For example, /P/ may be mispronounced as /pu/ so that the word *hope* is mispronounced as /H OW P U/ instead of /H OW P/.

To overcome this issue, an extra vowel phoneme is appended to the consonant if a consonant phoneme appears at the end of a word or it is followed by another consonant. For example, when the phonetic sequences of word *blog* are converted into Chinese Pinyin, the resulting Pinyin syllable sequences before and after considering the effect of L1 transfer are:

/B  L AA G / (English)
/b u l ao g / (direct mapping)
/b u l ao g e/ (considering effect of L1 transfer).

By applying the above rule, the phonological representations of English words that spoken by native Mandarin speakers are expected to be more accurate.

To recruit more English words into the Mandarin lexicon, a set of mapping rules on 9 English consonant phonemes were learned from a development set as described in Section 3.2. The rules are shown in Table 3. According to their corresponding phonemes in Chinese Pinyin, different vowels are appended by considering the articulatory factors. For example, /T/, /D/, /K/, /G/ are followed by vowel 'e'; /S/ and /Z/ are followed by vowel 'i'; Phonemes /P/, /B/ and /F/ are followed by vowel 'u'.

Table 3: *Phonemes mapping by considering the effect of negative L1 transfer.*

| English phonemes | Chinese Pinyin syllables |
|---|---|
| T, D, K, G | te, de, ke, ge |
| P, B | pu, bu |
| S, Z | si, zi |
| F | fu |
| M | mu |

*2.3.2. Nasal sounds not followed by vowels*

The 3 nasal phonemes /M/, /N/ and /NG/ in English are similar to the nasal phonemes in Mandarin, except phoneme 'm' never appears at the end of a proper Pinyin syllable. For example, the English syllable structure (C)(G)V(N), is similar to the combined (C)(G)V(N) structure in Chinese, e.g. *long* is a valid word in English as well as a valid Pinyin syllable. However, 'm' normally does not occur at the end of Pinyin syllable while many English words end with 'm' sound, e.g. *room*, *chrome*, etc. To address the issue of 'm', a vowel phoneme 'u' is proposed to append with 'm' at the end of a word. For example, when the phonetic sequence of word *chrome* is converted into Chinese Pinyin, the resulting sequences are:

/K  R AA M / (English)
/k  r ao m / (direct mapping)
/k e r ao m u / (considering effect of L1 transfer).

Table 3 summarized all the rules observed from the development set, and they can be combined with Table 2. To verify their effectiveness, the rules were applied on the ASR task of recognizing Mandarin utterances which contain English words. The details of the experiments are presented in the next section.

## 3. Experimental Setup

To validate the effectiveness of the proposed phonemes mapping rules, speech recognition experiments on mixed Mandarin and English utterances were conducted. A Large Vocabulary Continuous Speech Recognition (LVCSR) system was trained using Mandarin-only data and a pre-defined Mandarin lexicon. The phoneme mapping rules were applied to augment the lexicon with common English words, hence the ASR system became capable to recognize Mandarin utterances that contain those English words.

### 3.1. Mandarin LVCSR system

The Mandarin acoustic model training data was collected from smart phone users in Mainland China. The speakers were from a broad age range and gender balanced. The speech data was recorded in both quiet and noisy environment. It consisted of both read style and spontaneous style speech that recorded from different mobile operating systems such as iOS, Android and others. The data was manually transcribed. The total duration of the data is about 1700 hours, and the sampling rate is 16KHz.

The acoustic feature consists of 13 dimensional MFCC feature, 1 dimensional tone feature, and their derived deltas, acceleration and third-order deltas, resulting in 56 dimensions. The acoustic model contains 175 tonal monophones, and the context-dependent triphones are modeled by 8534 tied states. The final model was trained with deep neural network [21] on top of the model trained using maximum mutual information technique [22]. The Kaldi toolkit [23] was used to build the acoustic model.

The text data used in language model training was from 3 sources: i) text transcription of the acoustic model training set; ii) Chinese website data; iii) Specific domain data such as Sina weibo, Chinese city/place names and addresses. All the text data was segmented according to a lexicon which consisted of about 140K Mandarin words and 26 English alphabets. Finally, a 5-gram LM was developed using the SRILM toolkit [24].

### 3.2. Development and test data

The development and test data was recorded under the same condition as the acoustic model training data. Both speakers and utterances did not overlap with the acoustic model training data. Table 4 shows the details of the development and test sets. In *Dev* set, each utterance contains at least one English word. The 200 English words occurred in *Dev* set were used to generate the phoneme mapping rules.

Table 4: *Description of development and test sets.*

|  | No. Utter. | No. English Words | Description |
|---|---|---|---|
| *Dev* | 20942 | 200 | For learning pronunciation rules |
| *Test-E1* | 20958 | 200 | Utterances with at least 1 English word occurred in *dev* set |
| *Test-E2* | 4321 | 99 | Utterances with at least 1 English word, and the English words never occur in *dev* set |
| *Test-C* | 7999 | 0 | Utterances without English words |

The 3 test sets, namely *Test-E1*, *Test-E2* and *Test-C*, were used to validate the ASR performance. Each utterance in *Test-E1* and *Test-E2* contains at least 1 English word. The 200 English words occurred in *Test-E1* were as the same as *Dev* set, while the 99 English words in *Test-E2* were never observed in *Dev*. To study the effect of the augmented lexicon on Mandarin-only speech, *Test-C* set does not contain English words.

The English words found in the above data sets fall into the following two groups: proper names such as *iPad* and *photoshop*, and common words used in oral conversation such as *bye* and *hello*.

### 3.3. Incorporating foreign words in the LM

To elevate the LVCSR system to be able to recognize English words, the English words appeared in the development and test sets were added into the language model by updating the unigram statistics. Each English word was assigned a unified occurrence probability. After adding the English words, the language model was normalized and the the probabilities of the English words were tuned to maximize the perplexity of the transcription of the *Dev* set.

## 4. Experimental Results

The acoustic model described in Section 3.1 and the language model mentioned in Section 3.3 were used for all the experiments. The ASR performance using 3 different lexicons were compared: the original lexicon $L_o$, the augmented lexicon $L_d$ which derived from direct mapping rules in Section 2.2, and the further augmented lexicon $L_t$ which obtained by considering the effect of L1 transfer. The lexicon $L_t$ was extended from $L_d$ by adding extra pronunciation entries to the English words which are qualified for the mapping rules in Section 2.3.

The experimental results are shown in Table 5. The ASR performance is reported in character error rate (CER). In the scoring process, the Chinese words are separated into characters and each English word is considered as one character.

Table 5 shows that the ASR performance on *Test-E1* and *Test-E2* is improved when the pronunciations of English words are added into the lexicon. The performance using $L_t$ outperforms $L_d$. The results show that incorporating the knowledge of L1 transfer benefits foreign words pronunciation modeling. Furthermore, the improved performance on the open set *Test-E2* indicates that the proposed phoneme mapping rules are generalized and they can be directly applied to unseen words.

Table 5: *Performance of mixed Mandarin and English recognition using different lexicons.*

| CER(%) Lexicon | Test-E1 | Test-E2 | Test-C |
|---|---|---|---|
| $L_o$ | 33.33 | 43.14 | 10.75 |
| $L_d$ | 31.22 | 40.97 | 10.85 |
| $L_t$ | 30.72 | 39.98 | 10.87 |

Moreover, Table 5 indicates that only marginal changes in ASR performance on Mandarin-only set *Test-C* by using different lexicons. The findings verify that the proposed rule-based approach is effective in recognizing English words without harming the performance of Mandarin-only recognition.

To further analyze the experimental results, we break down the results into Mandarin only and English only sets, as shown in Table 6. The recognition performance of English words is measured by word error rate (WER). It is observed that the ASR performance of both Mandarin characters and English words are improved by using $L_t$ instead of $L_d$. The results are consistent for both *Test-E1* and *Test-E2* sets. It verifies that the pronunciation modeling is crucial in code-mixing speech data and the accurate pronunciation lexicon improves the recognition accuracy of both native and foreign words. The reported English WER in Table 6 is generally much higher than the Mandarin CER since the English words recognition suffers from high substitution error. It might be due to the weakness of the LM in modeling English words, as described in Section 3.3.

Table 6: *Breakdown of speech recognition results with different lexicons.*

|  | Mandarin CER (%) | | English WER (%) | |
|---|---|---|---|---|
|  | $L_d$ | $L_t$ | $L_d$ | $L_t$ |
| *Test-E1* | 25.21 | 24.83 | 88.91 | 87.26 |
| *Test-E2* | 32.88 | 32.02 | 93.61 | 91.77 |

Besides the above findings in the ASR performance, the following observation is obtained by investigating the decoding hypothesis: monosyllabic words are easier to be mis-recognized than multisyllabic words. For instance, most of the instances of words such as *high* and *day* were mis-recognized, while words such as *goodbye* and *wifi* had higher recognition accuracy. The observation can be interpreted using the fact that Mandarin follows a monosyllabic structure, and monosyllabic English words might be recognized as Mandarin characters.

## 5. Conclusion and Future Work

In this paper, the effect of language transfer is examined to derive the phonetic mapping rules to model pronunciation of English words using Mandarin phonemes. The ASR performance on mixed Mandarin and English utterances is improved by using lexicon mapping rules that considering the effect of language transfer. The proposed rules are generalized and it can be directly applied to unseen English words.

In the future work, we would like to study the way to combine the proposed rule-based method with the data-driven methods. As the language transfer rule is not able to cover all the pronunciation variations in human speech, we believe the data-driven method helps in providing complementary information for modeling the variation of pronunciations.

# 6. References

[1] S. Fitt, "The pronunciation of unfamiliar native and non-native town names," in *Proc. Eurospeech '95*, Madrid, Spain, Sep. 1995, pp. 2227–2230.

[2] F. Stouten and J.-P. Martens, "Recognition of foreign names spoken by native speakers," in *Proc. Interspeech '07*, Antwerp, Belgium, Aug. 2007, pp. 2744–2747.

[3] J. Y. C. Chan, H. Cao, P. C. Ching, and T. Lee, "Automatic recognition of cantonese-english code-mixing speech," *Computational Linguistics and Chinese Language Processing*, vol. 14, pp. 281–304, 2009.

[4] G. Stemmer, E. Noth, and H. Niemann, "Acoustic modeling of foreign words in a german speech recognition system," in *Proc. Eurospeech '01*, Aalborg, Denmark, Sep. 2001, pp. 2745–2748.

[5] Y. Li and P. Fung, "Improved mixed language speech recognition using asymmetric acoustic model and language model with code-switch inversion constraints," in *Proc. ICASSP '13*, Vancouver, Canada, May 2013.

[6] R. Tong, B. P. Lim, N. F. Chen, B. Ma, and H. Li, "Subspace gaussian mixture model for computer-assisted language learning," in *Proc. ICASSP '14*, Florence, Italy, May 2014.

[7] L. Lu, A. Ghoshal, and S. Renals, "Automatic data-driven pronunciation lexicon for large vocabulary speech recognition," in *Proc. ASRU 2013*, Olomouc, Czech Republic, Dec. 2013, pp. 374–379.

[8] I. McGraw, I. Badr, and J. Glass, "Learning lexicons from speech using a pronunciation mixture model," *IEEE Trans. Audio, Speech, and Language Processing*, vol. 21, no. 2, pp. 357–366, Feb. 2013.

[9] F. Beaufays, A. Sankar, S. Williams, and M. Weintraub, "Learning name pronunciations in automatic speech recognition systems," in *Proc. 15th IEEE International Conference on Tools with Artificial Intelligence*, California, USA, Nov. 2003, pp. 233–240.

[10] S. Schaden, "Rule-based lexical modelling of foreign-accented pronunciation variants," in *Proc. EACL 2003*, Budapest, Hungary, Apr. 2003, pp. 159–162.

[11] S. Zhang, Q. Shi, and Y. Qin, "Modeling syllable-based pronunciation variation for accented mandarin speech recognition," in *Proc. 2010 International Conference on Pattern Recognition (ICPR)*, Istanbul, Turkey, Aug. 2010, pp. 1606–1609.

[12] T. Modipa and M. H. Davel, "Pronunciation modelling of foreign words for sepedi ASR," in *Proc. 21st Annual Symposium of the Pattern Recognition Association of South Africa (PRASA)*, Stellenbosch, South Africa, Nov. 2010, pp. 185–189.

[13] C. B. Chang and A. Mishler, "Evidence for language transfer leading to a perceptual advantage for non-native listeners," *Journal of the Acoustical Society of America*, vol. 132, no. 4, pp. 2700–2710, 2012.

[14] T. Odlin, *Language Transfer*, 1st ed. Cambridge: Cambridge University Press, 1989.

[15] C. T. Best, "A direct realist view of cross-language speech perception," in *Speech perception and linguistic experience: Theoretical and methodological issues*, W. Strange, Ed. Baltimore: York Press, 1995, pp. 171–204.

[16] J. Flege, "Second language speech learning: Theory, findings and problems," in *Speech perception and linguistic experience: Theoretical and methodological issues*, W. Strange, Ed. Baltimore: York Press, 1995, pp. 233–277.

[17] D. I. Slobin, *The crosslinguistic study of language acquisition*. Hillsdale, NJ: Erlbaum, 1992, vol. 3.

[18] *CMU pronouncing dictionary. [Online] Available: http://www.speech.cs.cmu.edu/cgi-bin/cmudict*, Carnegie Mellon University.

[19] B. Ma and Q. Huo, "Benchmark results of triphone-based acoustic modeling on HKU96 and HKU99 putonghua corpora," in *Proc. ISCSLP 2000*, Beijing, China, Oct. 2000.

[20] R. Miao, "Loanword adaptation in mandarin chinese: Perceptual, phonological and sociolinguistic factors," Ph.D. dissertation, Stony Brook University, 2005.

[21] G. E. Dahl, D. Yu, L. Deng, and A. Acero, "Context-dependent pre-trained deep neural networks for large-vocabulary speech recognition," *IEEE Trans. Audio, Speech, and Language Processing*, vol. 20, no. 1, pp. 30–42, Jan. 2012.

[22] D. Povey, "Discriminative training for large vocabulary speech recognition," Ph.D. dissertation, Cambridge University Engineering Dept, 2003.

[23] D. Povey *et al.*, "The kaldi speech recognition toolkit," in *Proc. ASRU 2011*, Hawaii, USA, Dec. 2011.

[24] A. Stolcke, "SRILM - an extensible language modeling toolkit," in *Proc. ICSLP '02*, Denver, USA, Sep. 2002.